\documentclass{article}

\usepackage{arxiv}

\usepackage[utf8]{inputenc} 
\usepackage[T1]{fontenc}    
\usepackage{hyperref}       
\usepackage{url}            
\usepackage{booktabs}       
\usepackage{amsfonts}       
\usepackage{nicefrac}       
\usepackage{microtype}      
\usepackage{lipsum}

\usepackage{amsmath}
\usepackage{graphicx}
\usepackage{multirow}
\usepackage{subfig}

\usepackage{url}
\title{Multi-sense Definition Modeling using Word Sense Decompositions}

\author{
  Ruimin Zhu\\
  Department of Statistics\\
  Northwestern University\\
  \texttt{ruiminzhu2014@u.northwestern.edu} \\
  \And
  Thanapon Noraset\\
  Faculty of Information and Communication Technology\\
  Mahidol University\\
  \texttt{thanapon.nor@mahidol.edu} \\
  \And
  Alisa Liu\\
  Department of Computer Science\\
  Northwestern University\\
  \texttt{alisa@u.northwestern.edu}\\
  \And
  Wenxin Jiang\\
  Department of Statistics\\
  Northwestern University\\
  \texttt{wjiang@northwestern.edu}\\
  \And
  Doug Downey\\
  Department of Computer Science\\
  Northwestern University\\
  \texttt{ddowney@eecs.northwestern.edu}
}

\begin{document}
\maketitle

\begin{abstract}
Word embeddings capture syntactic and semantic information about words. Definition modeling aims to make the semantic content in each embedding explicit, by outputting a natural language definition based on the embedding. However, existing definition models are limited in their ability to generate accurate definitions for different senses of the same word.  In this paper, we introduce a new method that enables definition modeling for multiple senses. We show how a Gumble-Softmax approach outperforms baselines at matching sense-specific embeddings to definitions during training. In experiments, our multi-sense definition model improves recall over a state-of-the-art single-sense definition model by a factor of three, without harming precision.
\end{abstract}

\section{Introduction}
\label{sec:w2mdef_intro}
Distributed representations of words form a foundation for many of today's NLP systems. Word embeddings can be learned from raw text using a variety of techniques~\cite{turian_word_2010,mikolov_efficient_2013,pennington_glove_2014,yogatama_learning_2015}, and are used to represent words in NLP systems~\cite{socher2013recursive,cho_learning_2014,karpathy_deep_2014,xiong2016dynamic}. Words can have multiple meanings, and recent methods consider representing each word by multiple embeddings that represent its different senses~\cite{neelakantan2015efficient,tian2014probabilistic,pilehvar2016conflated,johansson2015embedding,arora_embdecomposition_2016}.

While the syntax and semantics of embeddings can be inspected indirectly through word similarity or analogy tasks, the recently introduced {\em definition modeling} task makes the semantics captured by an embedding explicit, by generating a natural language definition of a word in terms of its embedding \cite{nor_definition_2016}.  But, these methods cannot be applied to model multiple senses without using additional input, such as ground-truth example usage of the word sense being defined \cite{ni2017learning,gadetsky2018conditional}.

In this paper,  we investigate how to model multiple definitions for word from a {\em mixed-sense} embedding, in which the multiple senses of each word are initially mixed into a single embedding vector. Unlike previous work, we focus on extracting individual senses from a single embedding of polysemous words without example usage. Mixed-sense embedding is more commonly used in practice (e.g., the  Word2Vec \cite{mikolov_efficient_2013} and GloVe \cite{pennington_glove_2014} embedding methods), but also presents a challenge. In particular, since a word can have many definitions and many sense embeddings, it is {\em a priori} unknown which of the sense embeddings corresponds to which of the word's definitions. To address this challenge, we first extract sense embeddings ({\em atoms}) from a set of word embeddings using a recent multi-sense embedding approach \cite{arora_embdecomposition_2016}. And then, we explore two approaches including a heuristic that matches the sense embeddings to definitions before training and a Gumbel-Softmax (GS)~\cite{jang_gs_2016} approach that jointly selects assignments of definitions while training the definition model.

We compare and analyze different approaches of modeling multi-sense definitions with an emphasis on a workload of polysemous words. Our comparisons include both manual and automated evaluation. We find that our multi-sense models output equal or better quality definitions than the state-of-the-art model, but cover more distinct word senses. In addition, we find that the Gumbel-Softmax approach works somewhat better than the heuristic approach. Finally, our experiments also show that definition modeling remains a challenging task, especially for polysemous words, and an error analysis reveals several areas for improvement in future work.

\section{Previous Work}
Despite the success of single-sense word embeddings in modeling language, they suffer from an inability to discriminate among sub-senses of a word since each word is represented by a single vector. To address this issue, several models have been introduced for representing a word with multiple embeddings, one for each sense. Examples that learn multi-sense embeddings from text include the Multiple-Sense Skip-Gram Model of Neelakantan et al.~\cite{neelakantan2015efficient}, the finite mixture model of word embeddings introduced by Tian et al.~\cite{tian2014probabilistic}, the Topical Word Embeddings (TWE) proposed by Liu et al.~\cite{liu2015topical} and its extensions including Neural Tensor Skip-Gram (NTSG)~\cite{liu2015learning} and MSWE~\cite{nguyen2017mixture}.  

Another paradigm focuses on de-conflating existing single-sense word embeddings to obtain sense representations. Pilehvar and Collier~\cite{pilehvar2016conflated} plug word embedding vectors into the Personalized Page Rank algorithm~\cite{haveliwala2002topic} to learn sense embeddings. Jauhar et al.~\cite{jauhar2015ontologically} propose general approaches for generating sense-specific word embeddings that are grounded in an ontology. Johansson and Pina~\cite{johansson2015embedding} decompose word embeddings into a combination of its sense embeddings under the constraint that sense embeddings be close to their neighbors in the semantic network. Arora et al.~\cite{arora_embdecomposition_2016} show that word embeddings can be decomposed to sparse linear combinations of subsenses, or so-called \textit{atoms}.  We adopt the Arora et al. approach in this paper.

Definition modeling is the task of generating a natural language definition for a given word and its embedding~\cite{nor_definition_2016}.  Dictionary definitions tend to repeatedly utilize certain constructions to reflect semantics~\cite{markowitz1986semantically}, making them amenable to automated generation in many cases, provided that the semantics captured by an embedding is sufficiently accurate and comprehensive. We extend the definition models of~\cite{nor_definition_2016} to handle multiple senses, and show that utilizing multi-sense embeddings allows us to substantially improve recall over the models in that work.

Recent work from Ni and Wang~\cite{ni2017learning} generates different definitions for different senses of a word as well.  However, unlike our work they do not take multi-sense embeddings as input and they do not solve the problem of matching sense embeddings to dictionary definitions -- instead, their method requires an example usage in context of each word sense to be defined. Gadetsky et al.~\cite{gadetsky2018conditional} also generate sense-specific definitions by providing the model with word context, which is used to disambiguate multi-embeddings learned from an Adaptive Skip Gram model~\cite{bartunov2016breaking}, or to select components from a single-sense embedding for the word. Also, Yang, et al.,~\cite{yang2019incorporating} incorporated sememes, minimum semantic units, in their Chinese definition modeling task to generate sense specific definitions. In contrast to their work, our approach can be applied to a multi-sense embedding directly, and does not require ground truth example usage of each word as input.

\section{Multi-sense Embeddings}
\label{sec:w2mdef_atoms}
In this section, we describe the sense decomposition algorithm introduced by Arora et al.~\cite{arora_embdecomposition_2016} to obtain multi-sense embeddings from single sense word embeddings. In principle, the multi-sense definition modeling approach we will introduce can be applied to any of the multi-sense approaches discussed above. We choose Arora et al.'s~\cite{arora_embdecomposition_2016} approach due to its demonstrated effectiveness, and ease of implementation. The method takes a set of pre-trained single-sense word embeddings as input, and decomposes each word embedding into a sparse linear combination of {\em atoms} (multi-embeddings), each representing a different sense of the word:
\begin{equation}
v_w = \sum\limits_{j=1}^{m}\alpha_{w,j} A_j + \eta_w,
\end{equation}\label{equation:decomposition}
where $v_w$ is a single-sense word embedding for $w$, $A_j$ is an atom embedding, $\alpha_{w, j}$ is a coefficient giving the strength of the atom $A_j$ for the word $w$, and $\eta_w$ is a noise vector. In the sparse decomposition, the majority of the coefficients are zero. Atom embeddings are multi-embeddings that capture subsenses, and the atoms are shared across words.  For example:
\begin{multline*}
v(\text{cabinet}) = 0.93A_{344} + 0.47A_{1284} + 1.6A_{1520} + 0.47A_{2328} + 0.81A_{3092} + \eta_{cabinet}.
\end{multline*}
In the above, the single-sense embedding for {\tt cabinet} is decomposed into a linear combination of five atoms plus a noise term. To investigate the decomposition further, we can look at the nearest words to each atom in the embedding space, to get an idea of what each atom represents. For the above example, our inspection is shown in Table~\ref{table:atom-rep}.  From the nearest words, atom $A_{344}$ appears to reflect the furniture sense of the word -- we might expect that atom to also appear in the decomposition for the the word {\em dresser}.

\begin{table}[h!]
\begin{center}
\begin{tabular}{l|l}
\hline\hline
\bf atom & \bf nearest words \\
\hline
$A_{344}$ & closet, cupboard, drawers, ...\\
$A_{1284}$ & stoneware, china, dinnerware, ...\\
$A_{1520}$ & parliament, ministerial, ...\\
$A_{2328}$ & appointee, elected, appointed, ...\\
$A_{3092}$ & ministry, deputy, bureaucrat, ...\\
\hline
\end{tabular}
\end{center}
\caption{\label{table:atom-rep} A representation of each atom of the word {\tt cabinet} in terms of its nearest neighboring words (in cosine distance between a single-sense word embedding and an atom embedding).}
\end{table}

In our experiments, we use the pre-trained Word2Vec~\cite{mikolov_distributed_2013} embeddings learned from GoogleNews~\cite{mikolov2013efficient}. We run the decomposition algorithm on the embedding matrix of 50,000 common words, and set the sparsity parameter to five, i.e., each word can have at most five atoms.  The decomposition results in 4058 distinct atom embeddings. 

\section{Task and Data}
\label{sec:w2mdef_task}
In this section, we first define our task. We then describe our data consisting of dictionary definitions extracted from WordNet and the Oxford English Dictionary (OED). 

\subsection{Task definition}
In {\em Multi-sense Definition Modeling}, we are given a word, its single-sense embedding, and a set of atom embeddings that represent particular senses of the word. Our task is to maximize the probability of the set of natural language definitions of the word.  The task thus extends definition modeling~\cite{nor_definition_2016} to handle multi-sense embeddings.  Multi-sense embeddings are defined formally in Section \ref{sec:w2mdef_atoms}.

\subsection{Dictionary corpora}
\label{ssec:w2mdef_task_dict}

In our work, we extract data from two dictionaries: WordNet~\cite{wordnet,nltk} and OED\footnote{\url{https://developer.oxforddictionaries.com/}}.  We collect around 120,000 entries, where each entry is a tuple of a target word, its part of speech (POS), and its dictionary definition. The basic statistics of our definition corpus are shown in Table~\ref{table:corpus-statistics}. 

\begin{table}[t!]
\begin{center}
\begin{tabular}{l|l|l|l}
\hline\hline
\bf splits & \bf train & \bf valid & \bf test \\ 
\hline
\#words & 27006 & 1118 & 500 \\
\#entries & 111084 & 4745 & 4607\\
\#tokens & 1093130 & 45661 & 45707\\
average length & 9.8 & 9.6 & 9.9\\
\hline
\end{tabular}
\end{center}
\caption{\label{table:corpus-statistics} Basic statistics of the definition corpus used in this work. The three datasets are mutually exclusive.}
\end{table}

In this work, we also consider exploiting the part-of-speech of the word being defined. Part of speech is readily available from dictionaries.  Polysemous words often have different senses with different parts of speech, e.g. the word {\tt patient} can be a noun (``a person who needs medical care'') or an adjective (``being able to tolerate or endure unpleasantness''). Dictionary definitions usually exhibit certain patterns corresponding to parts of speech. For example, definitions with verb senses often follow a ``to $<$verb$>$'' structure, whereas adjective senses often begin with a gerund.  We hypothesize that modeling part of speech explicitly, and providing the model with the ground truth part of speech during training, helps the definition model identify the regularities and learn multiple senses more readily.

\section{Multi-sense Definition Models}
\label{sec:w2mdef_models}

We explore multiple models for the {\em Multi-sense Definition Modeling} task. All of the models are based on two building blocks: (1) A MATCH module used during the training phase to match each dictionary definition to the target word's atoms, using either heuristic or sampling-based approaches; and (2) a DEFINE module that generates definitions.

During the training stage, for each given dictionary definition of the target word, the MATCH module matches the definition to the target word's atoms. The DEFINE module is then trained using the matched atom embedding and the target word embedding as input, and the selected dictionary definition as the target output.  The MATCH module is an essential part of the system, because one of the fundamental challenges in multi-sense definition modeling is that the correspondence between atoms and training definitions is not known {\em a priori}.  We rely on MATCH to produce accurate training instances for DEFINE. In the testing stage, DEFINE is fed the target word embedding and each one of its atoms in turn for definition generation. 

The three multi-sense definition models we propose share the same DEFINE module architecture, and differ in MATCH. We begin our description of the models with DEFINE.

\subsection{The DEFINE module}
The core of DEFINE is a sequence-to-sequence block~\cite{cho_learning_2014,sutskever_sequence_2014}.  A shared two-layered LSTM first encodes the sequence: (\textit{target word}, $<$/s$>$), and then decodes to generate a definition for the target word.

To allow DEFINE to output a definition specific to a given atom, we allow the atom embedding affect the decoding process through a gated input~\cite{cho_learning_2014} similar to that used in a single-sense definition model. During the decoding stage, the decoder receives four gated inputs regarding the word and atom being defined: the word embedding, the matched atom embedding, a part of speech embedding, and the output of a character-level CNN~\cite{img_cnn} affix detector.  These gated inputs interact with LSTM hidden states as:
\begin{align}
z_t & = \sigma(W_z[v^*;h_t] + b_z),\\
r_t & = \sigma(W_r[v^*;h_t] + b_r),\\
\tilde{h}_t & = \tanh(W_h[r_t\odot v^*;h_t] + b_h),\\
o_t & = (1- z_t)\odot h_t + z_t\odot\tilde{h}_t,
\end{align}
where $[x;y]$ denotes vector concatenation, $\odot$ denotes element-wise multiplication, and $\sigma$ is the logistic sigmoid function. The variable  $v^*$ is a concatenation of the word embedding, atom embedding, part of speech embedding, and output of the CNN char-level affix detector, $h_t$ is the hidden state of the last LSTM layer at step $t$, and $o_t$ is the updated output.

\subsection{Heuristic matching}
For MATCH, our simplest approach adopts a heuristic for matching atoms to definitions based on the distance between the definition embedding and the atom embeddings. We notice that many words in a definition, such as function words, are not informative for identifying the word sense.  In fact, typically only one or two keywords in the definition are of essential importance.  Thus, we prune all function words and then for each atom define its distance to the definition as the sum of its cosine distances to the closest and second closest embeddings for words in the pruned definition. The atom with the smallest distance is taken as the matched atom. An example is given below.

{\tt cabinet}: a storage compartment for clothes and valuables (original)

{\tt cabinet}: storage compartment clothes valuables (function words removed)

\begin{table}[h!]
\centering
\begin{tabular}{l|l|l|l|l|l}
\hline\hline
 & storage & compartment & cloth & valuable & distance \\
\hline
$A_{344}$ & 0.376 & 0.530 & 0.206 & 0.090 & \bf{0.906} = 0.376 + 0.530\\
$A_{1284}$ & 0.087 & 0.176 & 0.305 & 0.093 & 0.481 = 0.305 + 0.176\\
$A_{1520}$ & 0.028 & 0.120 & 0.084 & -0.040 & 0.204 = 0.120 + 0.084\\
$A_{3092}$ & 0.042 & 0.050 & 0.028 & -0.015 & 0.092 = .050 + 0.042\\
\hline
\end{tabular}
\caption{Heuristic atom matching example. {\tt cabinet}: a storage compartment for clothes and valuables is matched to atom $A_{344}$.}
\label{table:heuristic_matching}
\end{table}

We refer to the model using this heuristic approach for matching as W2MDEF-HEU.

\subsection{Sampling-based matching}
W2MDEF-HEU performs a static, heuristic matching of definitions to atoms. We hypothesize that a more accurate approach should learn DEFINE and MATCH {\em jointly}, optimizing both during the training stage. We propose two joint sampling-based matching methods based on the Gumbel-Softmax technique~\cite{jang_gs_2016} which enables gradient flow in models where discrete variables must be sampled.

We first pass the dictionary definition to an encoder block, a two-layered LSTM, to encapsulate the semantics into a vector. Then, we calculate logits $\pi_i$  by multiplying the vector with each atom. Next, we sample a random atom $Z$ from a multinomial distribution where probabilities are characterized by the logits. However, the sampled index of the atom $Z$ is a discrete variable which blocks the gradient flow from  $Z$ to $\pi_i$, rendering gradient-based training impossible. The Gumbel-Softmax circumvents this problem by combining the Gumble-Max trick~\cite{gumbel1954statistical,maddison2014sampling} and a Softmax approximation to relax the one-hot $Z$ to be a continuous variable in a $k-1$ dimensional simplex:
\begin{align}
z_i &= \frac{\exp{(\log(\pi_i)+g_i)/\tau}}{\sum\limits_{j=1}^{k}\exp{(\log(\pi_j)+g_j)/\tau}},\\
g_i &\overset{\text{i.i.d}}{\sim} \text{Gumbel distribution}, i=1,\dots, k,
\end{align}
where $k$ is the number of atoms, and $\tau$ is a temperature hyperparameter controlling the sampling variance.

\begin{figure}[h!]
\centering
\includegraphics[scale=0.9]{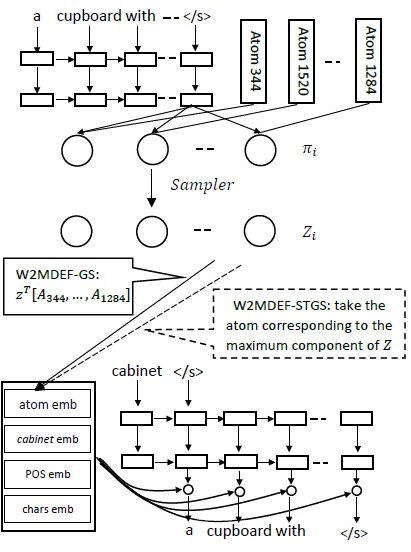}
\caption{\label{figure:architecture} Our W2MDEF-GS and W2MDEF-STGS architecture. The upper part is the MATCH module, and the bottom part is the DEFINE module. Note that in the DEFINE module, during the decoding stage, the LSTM is gated. W2MDEF-HEU also shares the same DEFINE module architecture.}
\end{figure}

Our model that uses the Gumbel-Softmax is referred to as W2MDEF-GS. In W2MDEF-GS, we feed the weighted atom embedding to DEFINE. We also experiment with a variant of GS sampling called a {\em straight-through} Gumbel-Softmax, where the atom corresponding to the maximum component of $Z$ is directly fed to DEFINE. We refer to the straight-through model as W2MDEF-STGS. Figure~\ref{figure:architecture} illustrates the MATCH and DEFINE modules of W2MDEF-GS and W2MDEF-STGS. 

\subsection{Part of Speech}
As we discussed in Section~\ref{ssec:w2mdef_task_dict}, we hypothesize that explicitly modeling part of speech might improve model performance. We add a POS input to the DEFINE module along with the atom embedding for gated input as shown in Figure~\ref{figure:architecture}. The ground truth POS is readily available during the training stage. However, it is unknown during the testing stage. We infer this information at testing time using the nearest-neighboring words of the atom. Often, the set of nearest words to an atom share a predominant POS. We take a majority vote of the parts of speech of the top 20 nearest neighboring words of the atom, and feed this POS to DEFINE.\footnote{Considering a different number of neighbors may result in higher accuracy, but we did not explore varying this parameter.}

\subsection{Dynamic regularization}
A common issue in neural language models is word and phrase repetition in model generated texts. We observed this phenomenon in {\em definition modeling} too, and adopt a recently proposed off-the-shelf regularizer (Reg)~\cite{noraset_controlling_2018} to mitigate the issue. During the training stage, the regularizer dynamically aligns the model-generated definitions' statistics, such as n-gram frequencies and repetition rates, to match those of the training corpus dictionary definitions.

\section{Experiments and Results}
\label{sec:w2mdef_experiments}
We now present the evaluation of our multi-sense definition models. For the purpose of testing model performance on the {\em Multi-sense Definition Modeling} task, we collect 500 common and highly polysemous words (not present in the training or validation set), such as {\tt tie}, {\tt capital}, and {\tt spring}, as our testing set. The average number of distinct senses of our testing set is 5, compared to the corpus average of 2.5.  We perform two types of evaluation: one using automated metrics, and another using manual labeling. 

\subsection{Experimental setup}
Our model architectures all use a 2-layered LSTM network with 300 units as the sequence-to-sequence (encoder-decoder) block in DEFINE. Both W2MDEF-GS and W2MDEF-STGS use another 2-layered LSTM network with 300 units to learn a vector representation for the dictionary definition in their MATCH modules. The word embeddings are fixed to the 300-dimensional Google News Word2Vec embeddings, and the atom embeddings are obtained using the semantic decomposition algorithm introduced by Arora et al.~\cite{arora_embdecomposition_2016}. The POS and character embeddings are 300 dimensional and initialized to uniformly randomly distributed small real numbers, and learned in the training stage. The affix detector uses a character-level CNN with kernels of length 2-6 and size {10, 30, 40, 40, 40} with a stride of 1. We apply a dropout~\cite{srivastava2014dropout} rate of 0.5 on LSTMs during the training stage to prevent models from overfitting. We use Adam~\cite{kingma_adam:_2014} to maximize the log-likelihood during the training stage. Starting with 0.001, the learning rate decays by a factor of 0.8 per epoch. Training is terminated after 2 consecutive epochs of no significant improvement, or after the learning rate is less than 1e-6. For W2MDEF-GS and W2MDEF-STGS, the softmax approximation temperature parameter in the Gumbel-Softmax sampler is initialized to 1.0 and anneals by a factor of 0.9 each epoch until it reaches 0.3. 

\subsection{Pruning}
During the testing stage, we iteratively feed the target word and one of its atoms to DEFINE. Ideally, the output definition for each atom should be distinct, so that they capture different senses of the target word.  However, the atom decomposition algorithm is imperfect, and the atoms do not always represent distinct senses.  In the illustration in  Section~\ref{sec:w2mdef_atoms}, for example, the five atoms of the word {\tt cabinet} are not mutually distinct. Atoms $A_{344}$ and $A_{1284}$ are both related to furniture, while atoms $A_{1520}$, $A_{2328}$, and $A_{3092}$ are all related to politics. This can lead to redundant model outputs. We add a heuristic merging step to alleviate this issue.

Specifically, after obtaining all definitions corresponding to different atoms, we calculate a similarity matrix using symmetrical BLEU score: 
\begin{equation}
d(o_i, o_j) = \frac{\text{BLEU}(o_i, o_j) + \text{BLEU}(o_j, o_i)}{2},
\end{equation}
where $o_i$ is the $i$-th model output. High symmetrical BLEU score indicates strong similarity between two model outputs. Outputs are merged into the same group whenever their symmetric BLEU score exceeds a threshold, set to 0.6 in our experiments.  The definition with the highest likelihood from each group is selected as a representative of that group and included in the final output set.

\subsection{Results}
Table~\ref{table:example_outputs} shows some selected outputs of our multi-sense definition models.
\label{ssec:results}
\begin{table}[h!]
\centering
\begin{tabular}{l|l|l}
\hline\hline
\bf word & \bf output & \bf model\\
\hline
\multirow{4}{*}{cabinet} & a small room , especially one used \\& for holding or storing things. & W2MDEF-GS (-Reg)\\
\cline{2-3}
& a legislative body of a country or other\\& group of people. & W2MDEF-GS (-Reg)\\
\hline
\multirow{2}{*}{squash} & a game played with a ball with a ball & W2MDEF-GS (-Reg, -POS)\\
\cline{2-3}
&  small fruit of the cabbage family. & W2MDEF-GS (-Reg, -POS)\\
\hline
\multirow{3}{*}{crane} & a tall tower with a wooden frame. & W2MDEF-GS (-Reg)\\
\cline{2-3}
& a tall wading bird with a long bill and\\&
long legs, typically having a long head \\& and long legs... & W2MDEF-GS\\
\hline
\end{tabular}
\caption{Selected examples of our multi-sense definition models' outputs.}
\label{table:example_outputs}
\end{table}

We compute BLEU of model outputs against ground-truth definitions to measure generation quality. When a target word has multiple ground-truth definitions, we take the maximum BLEU score against each of them. If a model has multiple outputs for a target word, we average each one's BLEU score first before averaging across all target words. However, BLEU does not reflect how many distinct senses of the target word are captured by model outputs. For example, consider a target word with four pairwise disjoint definitions $\{A, B, C, D\}$, and assume that Model 1 outputs $\{A\}$ whereas Model 2 outputs $\{A, C\}$. In this case, both models will get the same BLEU score: a perfect 100\%. But, Model 2 is superior to Model 1 on our task, since it captures more senses. To address this, we design a {\em rBLEU} (reverse BLEU) metric that switches the roles of ground-truth definitions and model outputs.  That is, rBLEU computes BLEU score treating ground-truth definitions as hypotheses, and model outputs as references. For the above example, the reverse BLEU for Models 1 and 2 would be 25\% and 50\% respectively, correctly identifying Model 2's superior performance. Since BLEU and rBLEU resemble precision and recall, we also obtain the {\em fBLEU} score by taking the harmonic mean of BLEU and rBLEU. 

We added another two baselines, NE and RANDOM, for comparison. NE returns the dictionary definitions of the training word which has the closest embedding to the target word to be defined. RANDOM shuffles the mapping between target words and definitions during training, i.e., trains on an incorrect dictionary.

We also introduce two additional baselines to investigate a potential confounding factor in our experiments. W2DEF outputs only a single definition for a given target word, whereas the other models can output different numbers of definitions.  Because rBLEU takes a maximum over the output definitions, a model that outputs more definitions has the advantage of having more chances to match tokens in the ground-truth definitions.
To address this concern, we also add a comparison when the models are restricted to output an approximately equal number of outputs per target word. Our multi-sense definition models output about three definitions per word. Therefore, for W2DEF, we choose its top three outputs with the highest likelihood scores, and we name this model $\text{W2DEF}^*$. For NE, we randomly sample three definitions per word, and we name this model $\text{NE}^*$. Table~\ref{table:bleu} lists the comparison results. 

\begin{table}[h!]
\begin{center}
\begin{tabular}{l|l|l|l}
\hline\hline
\bf model & \bf BLEU & \bf rBLEU & \bf fBLEU \\
\hline
W2DEF & 0.380 & 0.176 & 0.241\\
$\text{W2DEF}^*$ & 0.382 & 0.201 & 0.263\\
NE & \bf 0.431 & 0.252 & 0.318\\
$\text{NE}^*$ & \bf 0.431 & 0.223 & 0.294\\
RANDOM & 0.390 & 0.280 & 0.326\\
W2MDEF-HEU & 0.387 & 0.276 & 0.322\\
W2MDEF-STGS & 0.395 & 0.257 & 0.311\\
W2MDEF-GS & 0.409 & \bf 0.313 & \bf 0.355\\
\hline
\end{tabular}
\end{center}
\caption{Comparison the baseline model W2DEF and our Multi-sense Definition Models using BLEU, reverse BLEU, and fBLEU on 500 highly polysemous words.}
\label{table:bleu}
\end{table}

The results in Table \ref{table:bleu} show that W2MDEF-GS outperforms the other methods in fBLEU.  Comparing the results with equalized number of definitions to those without, we see that the rBLEU metric is sensitive to the number of definitions output by each method. The models and baselines will naturally output different numbers of definitions, but if we equalize these numbers the new approach maintains an advantage over the baselines.  However, in general we find that the BLEU-based metrics are not reliable for our task. They are unable to appropriately reflect the difference between semantically correct definitions and purely random ones. From these metrics, it seems that W2DEF slightly underperforms the random definition baseline RANDOM, but in manual inspection W2DEF is much better than RANDOM. 

We therefore evaluate on manual labeling to provide a more reliable measure of model performance. Before labeling, the outputs from different models are shuffled so that annotators don't know which model is behind each output. We manually label model outputs as one of four categories: {\bf I} the output is correct; {\bf II} the output has {\em either} a syntax/fluency error or a semantic issue, but not both; {\bf III} the output has both a syntactic and semantic error but is not completely wrong; and {\bf IV} where the output is completely wrong. Ground truth definitions are obtained from WordNet and OED. When evaluating precision and recall, the four labeling categories are given scores 1.0, 0.6, 0.3, and 0.0 respectively. Table~\ref{table:evaluation} demonstrates how the manual evaluation is performed.

\begin{table*}[h!]
\centering
\begin{tabular}{p{1.5cm}|p{1.5cm}|p{1cm}|p{1cm}|p{9cm}}
\hline\hline
\multicolumn{5}{c}{target word: squash}\\
\hline
\multicolumn{5}{c}{model outputs} \\
\hline
semantic group & label & score & atom & output \\
\hline
1 & II & 0.6 & $A_{1174}$ & to strike as if with a blow. \\
3 & II & 0.6 & $A_{0187}$ & a game played with a ball, typically with a curved blade and a round handle. \\
2 & I & 1.0 & $A_{0164}$ & a round flowered plant in the cabbage family, native to the US and New Zealand. \\
\hline
\multicolumn{5}{c}{ground-truth}\\
\hline
semantic group & source & \multicolumn{3}{p{11.5cm}}{definition}\\
\hline
1 & WordNet & \multicolumn{3}{p{11.5cm}}{to compress with violence to make out of shape.}\\
1 & OED & \multicolumn{3}{p{11.5cm}}{crush or squeeze with force so that it becomes flat , soft , or out of shape.}\\
2 & WordNet & \multicolumn{3}{p{11.5cm}}{any of annual trailing plants grown for their fleshy edible fruits.}\\
2 & OED & \multicolumn{3}{p{11.5cm}}{an edible gourd, the flesh of which may be cooked and eaten as a vegetable.}\\
3 & OED & \multicolumn{3}{p{11.5cm}}{a game in which two players use rackets to hit a small, soft rubber ball against the walls of a closed court.}\\
3 & WordNet & \multicolumn{3}{p{11.5cm}}{a game played in an enclosed court by two or four players who strike the ball with long-handled rackets.}\\
4 & OED & \multicolumn{3}{p{11.5cm}}{a state of being squeezed or forced into a small or restricted space.}\\
\hline
\multicolumn{5}{c}{evaluation}\\
\hline
precision & \multicolumn{4}{l}{$\frac{0.6+0.6+1}{3} = 0.73.$}\\
recall & \multicolumn{4}{l}{$\frac{0.6+0.6+1}{4} = 0.55.$}\\
\hline
\end{tabular}
\caption{\label{table:evaluation} Precision and Recall evaluation. Outputs are labeled as one of the four labeling categories (see text). Precision is average score, whereas recall is the sum of the maximum score achieved within each ground truth sense, divided by number of distinct ground truth senses.}
\end{table*}

Since manual evaluation is expensive, we choose our best model W2MDEF-GS from Table~\ref{table:bleu}, and compare it against the previous state-of-the-art single-sense definition model, W2DEF \cite{nor_definition_2016}, on a data set of 180 common and highly-polysemous words.  The results are shown in Table~\ref{table:model-comparison-2}. Models can have multiple definitions for a target word, and each annotator has to label 2032 examples. On this data set, the average pairwise correlation among three annotators is 0.78, indicating strong consensus on models' performance. Note that, however, multi-sense definition modeling is an extremely challenging task. From our experience, W2MDEF-GS usually captures at most two to three unique senses of polysemous words.

\begin{table}[h!]
\begin{center}
\begin{tabular}{l|l|l}
\hline\hline
\bf model & \bf prec & \bf rec \\
\hline
W2DEF & 0.155 & 0.036 \\
W2MDEF-GS & {\bf 0.204} & {\bf 0.116} \\
W2MDEF-GS (-Reg) & 0.180 & 0.106 \\
W2MDEF-GS (-Reg, -POS)& 0.155 & 0.091 \\
\hline
\end{tabular}
\end{center}
\caption{\label{table:model-comparison-2} Variants of our best model, W2MDEF-GS, compared against W2DEF on 180 polysemous words.  W2MDEF-GS boosts recall over W2DEF by 3x.  The results show that including part-of-speech and dynamic regularization improves accuracy.}
\end{table}

\begin{figure}[h!]
    \centering
    \subfloat[Precision]{{\includegraphics[width=7.5cm]{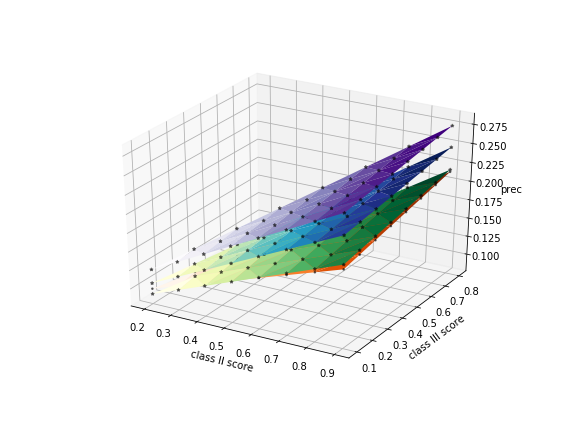}}}%
    \qquad
    \subfloat[Recall]{{\includegraphics[width=7.5cm]{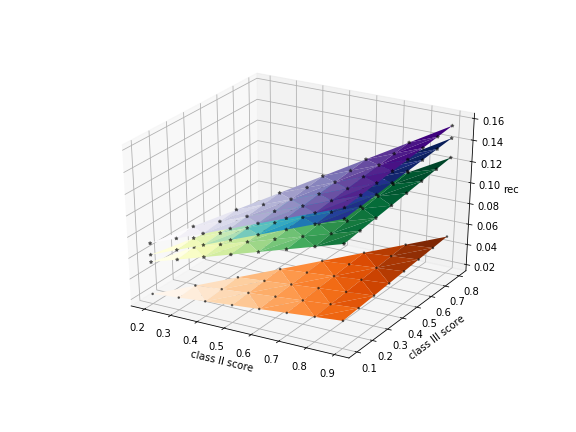}}}%
    \caption{Precision and recall plots under different scoring schemes with constraint: $1 = I > II > III > IV = 0$. Orange: W2DEF, Green: W2MDEF-GS (-Reg, -POS), Blue: W2MDEF-GS (-Reg), Purple: W2MDEF. Note that the monotonic relation is retained under different scoring schemes.}%
    \label{fig:sensitivity_analysis}
\end{figure}

Assigning fractional scores to the partially correct-answers is a subjective choice, so we verified that our results are not sensitive to this choice of scoring scheme.  Our sensitivity analysis revealed that for {\em any} choice of scores $a$ and $b$ for types II and type III errors with $b<a$ , the relative ranking of the methods in our experiments remained unchanged (Figure~\ref{fig:sensitivity_analysis}). Also, to ease interpretation of our results we provide a break down of error types. 

A break-down of error types is shown in Table~\ref{table:error-analysis}.  Compared to W2DEF, our best model W2MDEF-GS tends to generate fewer completely incorrect outputs, and has less redundancy\footnote{In W2DEF, we use slightly different heuristics for removing repetition from those in \cite{nor_definition_2016}, which allows slightly more repetition than the original model would have, but these differences are not significant in the final results.} which may be attributable to the dynamic regularizer (Reg). We notice that other than entirely wrong outputs, Under or Over-specification and Inaccurate modifier(s) are two major issues that limit definition modeling performance.  

\begin{table}[h!]
\centering
\begin{tabular}{p{2cm}|p{4.8cm}}
\hline\hline
W2DEF & W2MDEF-GS\\
\hline
\multicolumn{2}{l}{Completely Incorrect}\\
\multicolumn{2}{l}{stamp: a person's face.}\\
76.1\% & 71.1\%\\
\hline
\multicolumn{2}{l}{Redundancy and repetition}\\
\multicolumn{2}{l}{racket: a loud, loud, loud noise.}\\
7.0\% & 3.5\% \\
\hline
\multicolumn{2}{l}{Wrong POS}\\
\multicolumn{2}{l}{odd: to make a mystery or bewildering to.} \\
0.0\% & 0.8\% \\
\hline
\multicolumn{2}{l}{Close but inaccurate semantics}\\
\multicolumn{2}{l}{company: a person who sells goods.}\\
2.8\% & 4.0\% \\
\hline
\multicolumn{2}{l}{Under or Over-specified} \\
\multicolumn{2}{l}{pupil: a person who is a member of a school.} \\
6.0\% & 8.5\% \\
\hline
\multicolumn{2}{l}{Inaccurate modifier(s)}\\
\multicolumn{2}{l}{column: a short, legged essay or journal.}\\
6.5\% & 7.2\% \\
\hline
\multicolumn{2}{l}{Opposite}\\
\multicolumn{2}{l}{sanction: the act of restraining a punishment.}\\
0.6\% & 0.8\% \\
\hline
\multicolumn{2}{l}{Mixture of two or more subsenses}\\
\multicolumn{2}{l}{novel: a new or literature work.}\\
0.0\% & 1.0\% \\
\hline
\end{tabular}
\caption{\label{table:error-analysis} A breakdown of error types, and the percentage of each in our best model vs. the baseline. For each error type, we give one example. An output can have more than one error type.}
\end{table}

\section{Discussion}
\label{sec:w2mdef_discussion}
In this section, we present discussion on several components of the model as well as a quantitative analysis of model error.

\subsection{Matching}
\begin{figure}[h!]
\centering
\includegraphics[scale=0.45]{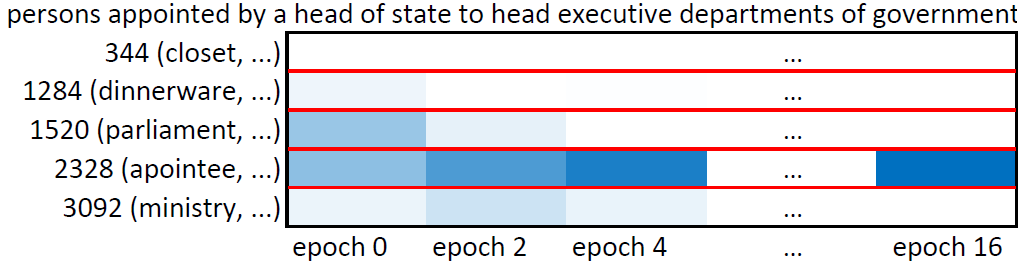}
\caption{\label{figure:atoms}Heat map of how weights of atoms of the target word {\em cabinet} update during the training stage. As training advances, weights gradually concentrate on a single atom.}
\end{figure}

Training a Multi-sense Definition model relies on accurate atom matching during the training stage.  Figure~\ref{figure:atoms} shows an example how the weights on atoms update during the training stage in W2MDEF-GS. The illustration shows that at the early stages of training, a definition is usually mapped to multiple atoms, and as training proceeds, the weights gradually concentrate on a single one.  Unlike the other two methods (W2MDEF-STGS and W2MDEF-HEU), W2MDEF-GS exposes this uncertainty to DEFINE in the form of a weighted average atom embedding.  The STGS approach by contrast always chooses the most likely atom, and the HEU model depends on making a good heuristic atom matching at the beginning.  We hypothesize that preserving atom uncertainty and exposing it to DEFINE early in training helps the Multi-sense Definition model achieve better atom matching.

\subsection{Gating}
\begin{figure}[h!]
\centering
\includegraphics[scale=0.25]{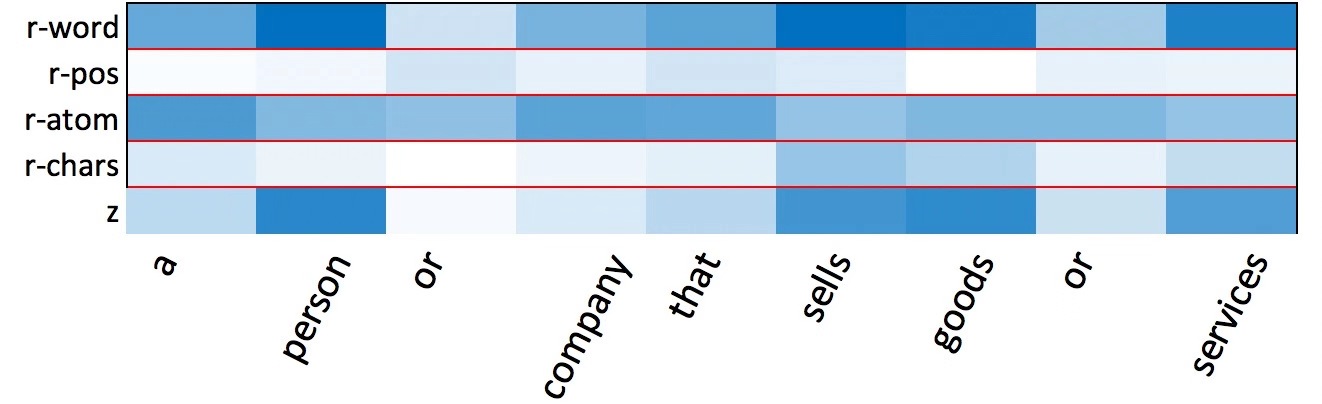}
\caption{\label{figure:gates}Gating of the target word {\em firm} during decoding stage. The ``r'' rows indicate the average reset gate value for each component---darker cells indicate that a component is more important for the given token.  The ``z'' row gives the average update gate value of the LSTM.}
\end{figure}

The gating mechanism in DEFINE coordinates how much information to take from each source, such as the LSTM hidden state, atom embedding, and character embedding, when generating definitions. Figure~\ref{figure:gates} shows one illustration of how the gates dynamically change during the decoding stage.  We observe that the target word embedding usually actively influences word generation throughout the decoding process, especially for non-function words. And as expected, the atom embedding can play an important role in distinguishing word senses. In defining the word ``firm'', for example, the first token (``a'') suggests a noun rather than an adjective sense, and the atom is active for this token. We also experimented turning off character embeddings and observed little impact on the model's performance.

\subsection{Quantitative analysis of error types}
In order to better understand the settings under which the W2MDEF-GS model succeeds and fails, we investigated whether certain attributes of words and atoms are predictive of model performance. The word attributes we considered included word frequency, the number of ground-truth definitions of the word, the semantic diversity of ground-truth definitions, and the word embedding norm.  Atom attributes included the atom weight after decomposition and the part of speech of the atom. We used logistic regression with these attributes to predict two different output variables: the individual error types (from Table \ref{table:error-analysis}), and the 0-1 manual evaluation score (from Section \ref{ssec:results}). For predicting the score, we trained logistic regression to minimize the cross-entropy between the model output and the score (i.e., we treated the non-0/1 score labels as probabilities). We performed 5-fold validation, where atoms belonging to the same word must always be in the same fold. 

We were unable to predict the individual error labels with accuracy above baseline, which suggests the attributes were not good predictors given the scale of data we had available, and demonstrates that definition generation is still a challenging problem. However, the score prediction model predicts the score with 0.48 loss, compared to the 0.53 baseline, using {\em only} atom weight as an attribute, which is a significant predictor with p-value $<$ 0.01. We speculate that this is because atoms with greater weight are more likely to represent more dominant senses that are easier to define.  In fact, the atoms with the top 10\% in weight have an average score of 0.35, substantially higher than the average score of 0.19 across all atoms.

\section{Conclusion}
\label{sec:w2mdef_conclusion}
In this work, we studied the {\em Multi-sense Definition Modeling} task. Our work takes a sense decomposition of pre-trained word embeddings and applies sequence-to-sequence neural nets to generate natural language definitions for each sense. We introduced novel approaches to match atoms to definitions during the training stage. Our best model, W2MDEF-GS, jointly trains a matcher and a definition generator using a Gumbel-Softmax technique. W2MDEF-GS substantially increases recall compared to existing definition modeling approaches.  Our error analysis identified several areas for improvement in the models. 

\section{Acknowledgments}

This work was supported in part by NSF Grant IIS-1351029 and the Allen Institute for Artificial Intelligence.  We thank Yiben Yang and the anonymous reviewers for their helpful feedback. 

\bibliography{references} 

\end{document}